\documentclass{article}

\usepackage{arxiv}

\usepackage[utf8]{inputenc} 
\usepackage[T1]{fontenc}    
\usepackage{hyperref}       
\usepackage{url}            
\usepackage{booktabs}       
\usepackage{amsfonts}       
\usepackage{nicefrac}       
\usepackage{microtype}      
\usepackage{todonotes}
\usepackage{graphicx}
\usepackage{lscape}
\usepackage{subcaption}
\usepackage{booktabs}
\usepackage{float}
\usepackage{subcaption}
\usepackage{framed,multirow}
\usepackage{setspace}
\usepackage{threeparttable}
\usepackage{tabularx}
\usepackage{amsmath}

\newcolumntype{L}[1]{>{\raggedright\arraybackslash}p{#1}} 
\newcolumntype{C}[1]{>{\centering\arraybackslash}p{#1}} 
\newcolumntype{R}[1]{>{\raggedleft\arraybackslash}p{#1}} 

\usepackage{amssymb}
\usepackage{latexsym}

\usepackage{xcolor}
\usepackage{todonotes}

\usepackage{hyperref}
\usepackage{booktabs}

\usepackage[official]{eurosym}
\usepackage{pdfpages}
\usepackage{stackengine}
\usepackage{algorithm}
\usepackage{algorithmic}

\definecolor{newcolor}{rgb}{.8,.349,.1}

\usepackage[official]{eurosym}

\usepackage{ mathrsfs }
\usepackage{pifont}

\definecolor{dkfzgreen}{rgb}{0,0.6,0.2}
\definecolor{dkfzyellow}{rgb}{1,0.839,0}
\definecolor{dkfzorange}{rgb}{1,0.361,0}
\newcommand{\cmark}{{\color{dkfzgreen} \Large{\ding{51}}}}%
\newcommand{\xmark}{{\color{dkfzorange} \Large{\ding{55}}}}%
\DeclareMathOperator*{\argmax}{arg\,max}

\title{How can we learn (more) from challenges? A statistical approach to driving future algorithm development.}

\author{
  \small Tobias Roß$^{1,2, }$\thanks{First authors contributed equally to this paper. Contact email address: \texttt{t.ross@dkfz-heidelberg.de}}\\
   \And
   \small Pierangela Bruno$^{1,4, *}$\\
   \And
    \small Annika Reinke$^{1,3,5}$\\
    \And
    \small Manuel Wiesenfarth$^{6}$\\
    \And
    \small Lisa Koeppel$^{7}$\\
    \And
    \small Peter M. Full$^{2,8}$\\
    \And
    \small Bünyamin Pekdemir$^{1}$\\
    \And
    \small Patrick Godau$^{1,5}$\\
    \And
    \small Darya Trofimova$^{1,11}$\\
    \And
    \small Fabian Isensee$^{3,8,11}$\\
    \And
    \small Sara Moccia$^{9}$\\
    \And
    \small Francesco Calimeri$^{4}$\\
    \And
    \small Beat P. Müller-Stich$^{10}$\\
    \And
    \small Annette Kopp-Schneider$^{6}$\\
    \And
    \small Lena Maier-Hein$^{1,2,5}$\\
}

\begin{document}
\maketitle

\begin{scriptsize} 
    $^1${Computer Assisted Medical Interventions (CAMI), German Cancer Research Center (DKFZ), Heidelberg, Germany}\\
    $^2${Medical Faculty, Heidelberg University, Heidelberg, Germany}\\
    $^3${HIP Helmholtz Imaging Platform, German Cancer Research Center (DKFZ), Heidelberg, Germany}\\
    $^4${Department of Mathematics and Computer Science, University of Calabria, Rende, Italy}\\
    $^5${Faculty of Mathematics and Computer Science, Heidelberg University, Germany}\\
    $^6${Division of Biostatistics, German Cancer Research Center (DKFZ), Heidelberg, Germany}\\
    $^7${Section Clinical Tropical Medicine, Heidelberg University, Heidelberg, Germany}\\
    $^8${Division of Medical Image Computing (MIC), German Cancer Research Center (DKFZ), Heidelberg, Germany}\\
    $^9${The BioRobotics Institute and Department of Excellence in Robotics and AI, Scuola Superiore Sant'Anna, Italy}\\
    $^{10}${Department for General, Visceral and Transplantation Surgery, Heidelberg University Hospital, Heidelberg, Germany}\\
    $^{11}${HIP Applied Computer Vision Lab, MIC, German Cancer Research Center (DKFZ), Heidelberg, Germany}\\
\end{scriptsize}

\newpage
\begin{abstract}
Challenges have become the state-of-the-art approach to benchmark image analysis algorithms in a comparative manner. While the validation on identical data sets was a great step forward, 
results analysis is often restricted to pure ranking tables, leaving relevant questions unanswered. Specifically, 
little effort has been put into the systematic investigation on what characterizes images in which state-of-the-art algorithms fail. To address this gap in the literature, we (1) present a statistical framework for learning from challenges and (2) instantiate it for the specific task of instrument instance segmentation in laparoscopic videos. Our framework relies on the semantic meta data annotation of images, which serves as foundation for a General Linear Mixed Models (GLMM) analysis. Based on 51,542 meta data annotations performed on 2,728 images, we applied our approach to the results of the Robust Medical Instrument Segmentation Challenge (ROBUST-MIS) challenge 2019 and revealed underexposure, motion and occlusion of instruments as well as the presence of smoke or other objects in the background as major sources of algorithm failure. Our subsequent method development, tailored to the specific remaining issues, yielded a deep learning model with state-of-the-art overall performance and specific strengths in the processing of images in which previous methods tended to fail. Due to the objectivity and generic applicability of our approach, it could become a valuable tool for validation in the field of medical image analysis and beyond.  and segmentation of small, crossing, moving and transparent instrument(s) (parts).
\end{abstract}


\newpage
\section{Introduction}
Comparative performance assessment of image analysis algorithms is typically performed by either reimplementing state-of-the-art methods or by international benchmarking competitions, so-called challenges~\cite{maier2018rankings}. As the re-implementation of other methods is prone to errors (e.g., errors in the implementation or suboptimal choice of hyperparameters) and time-consuming, challenges are nowadays the de facto standard for benchmarking new methods. 

To date, however, relatively little effort has been put into the systematic analysis of results, as summarized in~\cite{wiesenfarth2021methods}.

Specifically, most reports neglect a particularly relevant question for the medical domain:\\ 

\textit{What characterizes images on which algorithms fail?} \\

Or, more broadly speaking:\\

\textit{How can we learn from challenge results in a way that enables us to tailor future algorithm development to the specific remaining needs?}\\

\begin{figure*}
    \centering
    \includegraphics[width=1\linewidth]{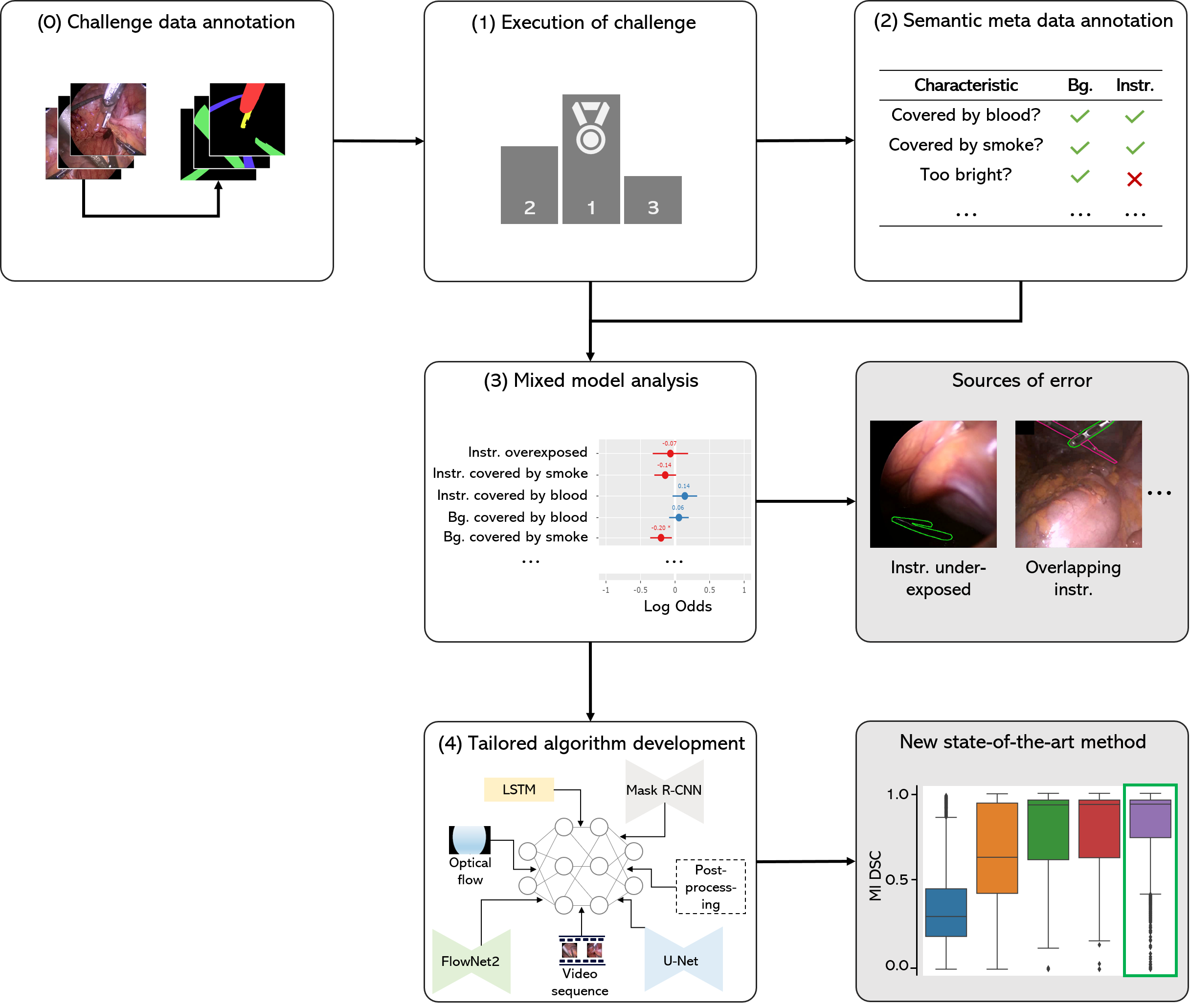}
    \caption{Overview of the statistical approach to learning from a challenge. Based on the annotated challenge data set (0) and the results of the given challenge (1, here: Robust-MIS challenge), a semantic annotation of the challenge's test cases (2) is performed. This serves as the basis for a (generalized) linear mixed model ((G)LMM) analysis (3) to identify major sources of algorithm failure and quantify the respective impact. The algorithm development is then tailored to the specific weaknesses (4) with the goal of enabling a new state-of-the-art performance. (Bg: Background; Instr: Instrument) }
    \label{fig:framework_overview}
\end{figure*}

\begin{figure}[tbp]
    \centering
    \begin{tabular}{cc}
    \includegraphics[width=.40\linewidth]{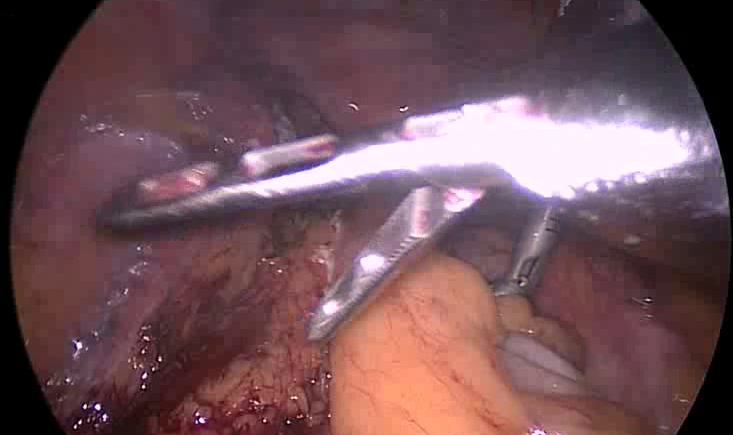}&
    \includegraphics[width=.40\linewidth]{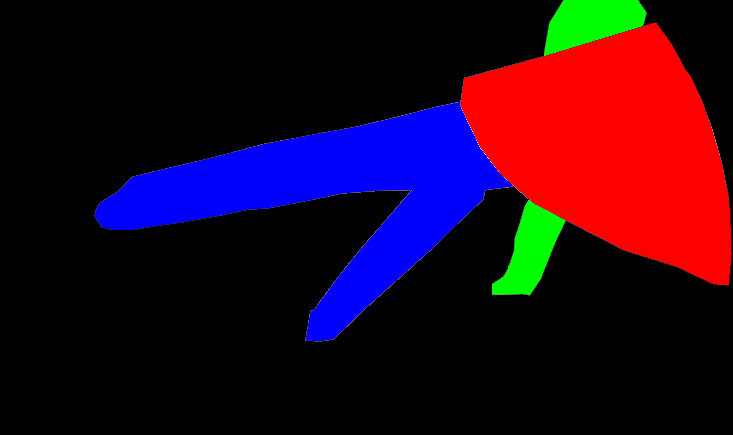}\\
    a) RGB image frame& 
    b) multi-instance segmentation
    \end{tabular}
    \caption{(a) Sample video frame with multiple overlapping instruments and (b) corresponding annotation (red: trocar, green/blue: graspers). Challenges include presence of specular reflections, noise, motion blur, varying illumination levels, overlapping instruments and instrument occlusions.}
    \label{fig:robust_mis_example_overlaps}
\end{figure}

Some challenge organizers have recognized this problem and carried out a laborious manual analysis, e.g., by reporting the best and worst cases based on the participants' performances and identifying a set of image characteristics that could lead to worsening or improving performance (i.e., over-/underexposed images)~\cite{ross2020robust, allan2021stereo, allan20202018, allan20192017}. However, this approach is rather subjective and does not allow for reliable quantification of the effects of different sources. Furthermore, it may be subject to confirmation bias. 
Given the lack of systematic analysis methodology, the contribution of this paper is threefold: 
\begin{enumerate}
    \item We present a statistical framework for learning from challenges, which focuses on the identification of sources for algorithm failure (Fig.~\ref{fig:framework_overview}). 
    \item To demonstrate potential benefit of the new concept, we apply it to the recently published challenge on multi-instance laparoscopic instrument segmentation ROBUST-MIS \footnote{Robust Medical Instrument Segmentation (RobustMIS) Challenge 2019, https://www.synapse.org/\#!Synapse:syn18779624/wiki/591266}~\cite{ross2020robust} (see Fig.~\ref{fig:robust_mis_example_overlaps}).
    \item We demonstrate that knowledge on the identified sources of error can help improve algorithm performance in common failure cases.
\end{enumerate}
Note that this approach was specifically designed for challenges but is similarly applicable to more basic validation studies, in which the performance of only a single algorithm is assessed.

The remainder of this paper is structured as follows: After presenting the related work in Sec.~\ref{sec:related_work}, we describe our framework for challenge analysis, its application to the task of multi-instance segmentation as well as our strength-weakness-driven algorithm development in Sec.~\ref{sec:methods}. The performed experiments and results are presented in Sec.~\ref{sec:experiments} and discussed in Sec.~\ref{sec:discussion}. Sec.~\ref{sec:conclusion} concludes this paper by summarizing our main findings.

\section{Related Work}\label{sec:related_work}

In this section, we present the related work on systematic challenge analysis in the field of biomedical image analysis as well as a brief summary on the state of the art in multi-instance medical instrument segmentation.

\subsection{Challenge analysis}
The literature on the analysis of image analysis challenges is extremely sparse, both within and outside the medical community. In fact, the meta science papers published to date have focused on ranking instabilities~\cite{maier2018rankings, reinke2018exploit}, standards~\cite{mendrik2019framework, maier2018rankings,maier2020bias} and challenge visualization~\cite{wiesenfarth2021methods}, while the topic of results analysis has been given extremely little attention. The closest work to ours was only recently published and presents a framework for visualizing challenge results in an uncertainty-aware manner~\cite{wiesenfarth2021methods}. While sources of error are not addressed within the framework, the paper provided an important motivation for our work: An analysis of numerous challenge reports in the field of biomedical image analysis revealed that a large number (66\% of those investigated) report only final ranks or aggregated performance measures \cite{wiesenfarth2021methods} without providing further analyses. This finding is in line with our more recent observations: Challenge reports often only provide a website with the rankings (e.g. MISAW\footnote{https://www.synapse.org/\#!Synapse:syn21776936/wiki/601705}, SurgVisDom\footnote{https://www.synapse.org/\#!Synapse:syn22083820/wiki/606329}, EndoVis-WorkFlowChallenge\footnote{https://endovissub2017-workflow.grand-challenge.org/PastChallenges/}), and corresponding publications concentrate on the presentation of aggregated metric values \cite{al2019cataracts, bodenstedt2018comparative}, visual examples \cite{allan20192017, allan2021stereo} or manual inspection of best/worst cases~\cite{ross2020robust}. In fact, we are not aware of any prior work on identifying sources of algorithm failure in a systematic manner.

\subsection{Multi-instance segmentation}
While the task of binary instrument segmentation received a lot of attention over the last couple of years such as~\cite{lee2019segmentation, shvets2018automatic, jin2019incorporating, allan2015image, garcia2016real}, literature on multi-instance segmentation in applications for minimally invasive surgeries is extremely sparse. To our knowledge, the only peer-reviewed work published independently of the ROBUST-MIS challenge~\cite{ross2020robust} (which this work is based on), was published by Shvets et al. \cite{shvets2018automatic}. Their work is on robotic instrument segmentation and features a comparatively simplistic data set with respect to image characteristics (e.g., blood, reflections). Hence, the methods competing in the ROBUST-MIS challenge can be regarded as representative for the state of the art in the field.

\section{Methods}
\label{sec:methods}
The following sections present the proposed framework for learning from challenges (Sec.~\ref{sec:methods:framework}), its instantiation in the ROBUST-MIS challenge (Sec.~\ref{sec:methods:instantiation}), as well as the proposed deep learning method resulting from problem-tailored algorithm development (Sec.~\ref{sec:methods:algorithm_development}).

\subsection{Framework for learning from challenges}
\label{sec:methods:framework}
This section introduces our concept for learning from challenges and details the underlying statistical approach.

\subsubsection{Concept overview}
We propose the following four-step procedure for learning from challenges.

\begin{enumerate}

\item \textbf{Hypothesis generation:} In an initial step, potential sources of algorithm failure are identified. These can relate to the image device (e.g. dirty endoscope lens), the imaging protocol (e.g. overexposure/underexposure), the handling of the equipment (e.g. motion blur), the target structure (e.g. crossing medical instruments in the case of ROBUST-MIS) and other application-specific features (e.g. smoke in the field of view of a laparoscope). To generate a list of image characteristics that may lead to poor performance, knowledge from the literature, expert knowledge, personal experience, as well as a manual analysis of the challenge results (as in~\cite{ross2020robust}) can be leveraged. 

\item \textbf{Semantic meta data annotation:} (Part of) the challenge test cases are then semantically annotated with these image characteristics. This can be done by domain experts, or by leveraging crowdsourcing, for example.

\item \textbf{Mixed model analysis:} The semantic labels on image characteristics along with the challenge results -- represented by (aggregated) metric results per test case -- are then leveraged to identify image characteristics leading to poor algorithm performance. To this end, a mixed effects model \cite{west2014linear} is set up in which the (possibly transformed) metric values embody the outcome variable and the image-specific information is integrated as explanatory variables. In other words, the performance of an algorithm on a given image is represented as a function of the meta information available for the image. The method is detailed in Sec.~\ref{sssec:MM-analysis}.

\item \textbf{Tailored algorithm development:} Based on the identification of those error sources that have the biggest effect on algorithm performance, algorithm development is tailored to the specific problems identified.
\end{enumerate}

Our approach to mixed model-based challenge analysis is detailed in Sec.~\ref{sssec:MM-analysis}, followed by an instantiation of this framework in the specific task of multi-instance instrument segmentation in laparoscopic video data in Sec.~\ref{sec:methods:instantiation}.

\begin{figure*}[tbp]
    \centering
    \includegraphics[width=0.8\linewidth]{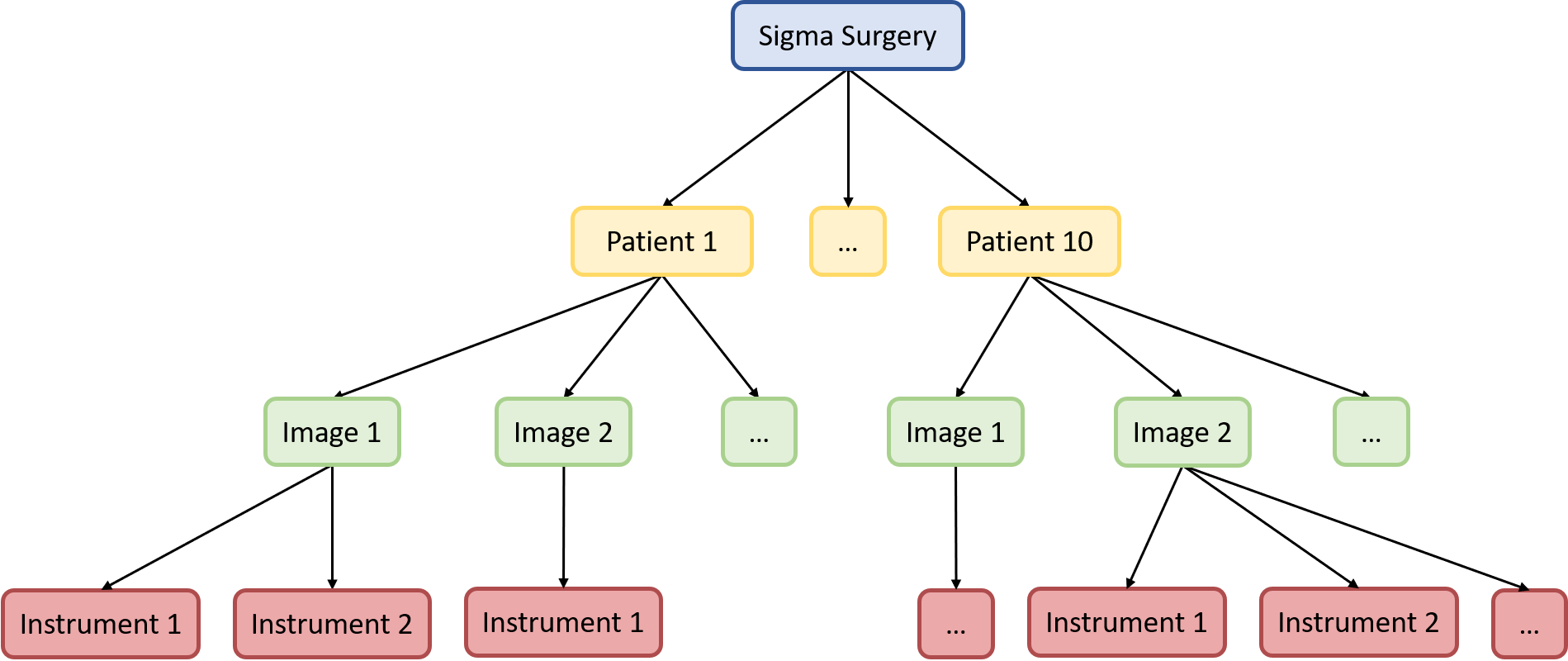}
    \caption{Hierarchical structure of the ROBUST-MIS 2019 data. The stage 3 test set comprises solely the Sigma surgery that was performed for ten patients. For each patient $p \in P$, a varying number of images are acquired. Every image $i \in I$ itself contains a varying number of instrument instances $J_i$.} 
    \label{fig:hierarchy}
\end{figure*}

\begin{figure*}[tbp]
    \centering
    \includegraphics[width=1\linewidth]{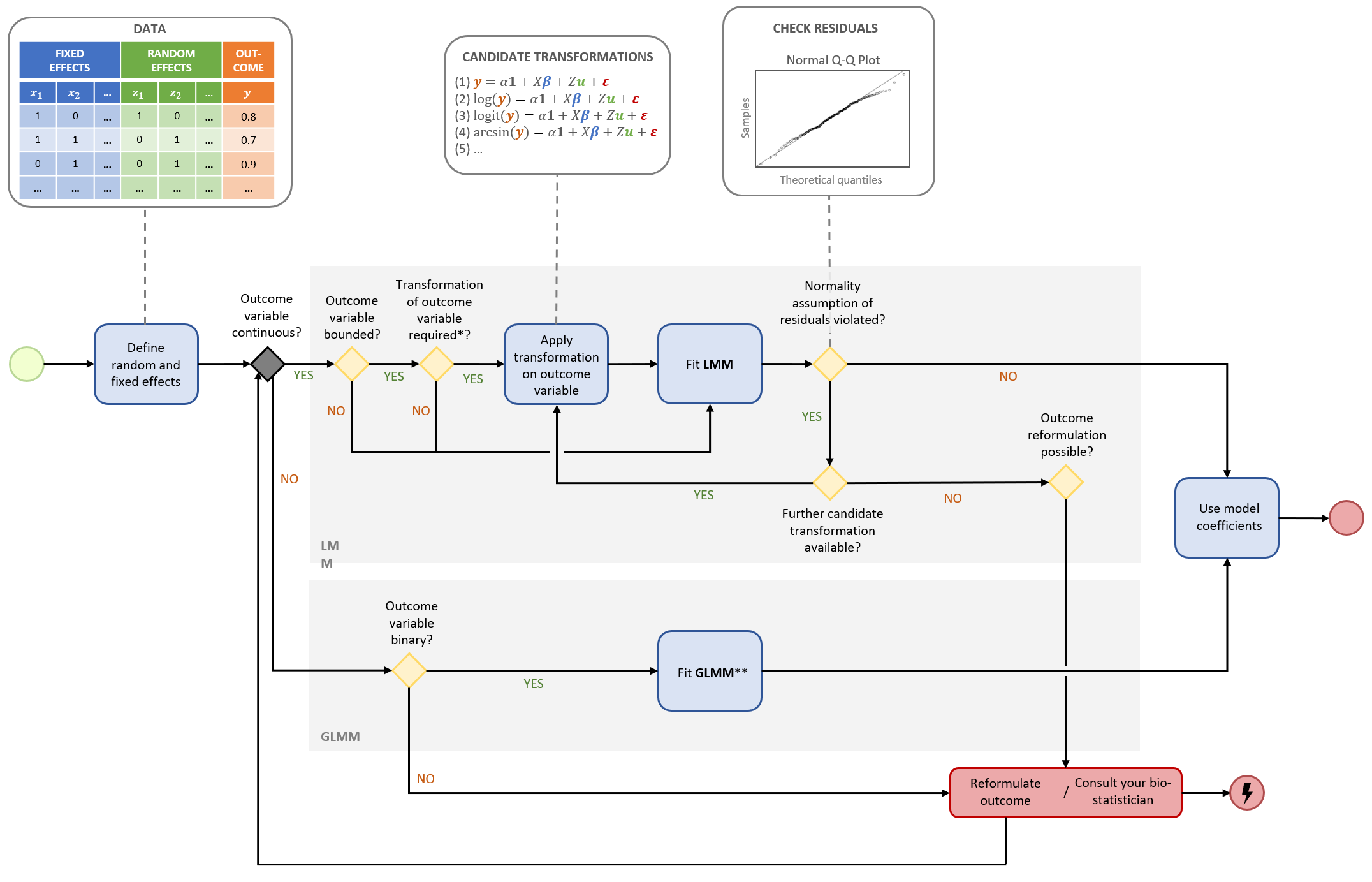}
    \caption{Mixed model-based statistical analysis for a challenge with a continuous metric value as outcome. Initially, random and fixed effects are defined. Image characteristics that have a potential influence on algorithm performance are assumed to be binary and are represented as fixed effects in the model. Other factors, reflecting the hierarchical structure of the acquired data (e.g. the patient/hospital/image frame identifier) are represented as random effects. Depending on the distribution of the outcome, either a Linear Mixed Model (LMM) or a Generalized LMM (GLMM) for binary outcome is the model of choice. In the case of the LMMs, a transformation of the outcome may be required before the model is fitted to avoid violation of the normality assumption. Further details of the workflow are provided in Sec. \ref{sssec:MM-analysis}. and Sec. \ref{sec:methods:instantiation}.\\
    \scriptsize *Note that metric values may be bounded in theory but appear normally distributed in the specific data set. In this case, no transformation is needed. \\
    \scriptsize **Note that a GLMM for binary outcome is a mixed effect logistic regression model.} 
    \label{fig:statistical_modeling}
\end{figure*}

\subsubsection{Mixed model analysis}
\label{sssec:MM-analysis}
As summarized in the previous section, the goal of the statistical analysis is to leverage the semantic meta data annotation to identify image characteristics leading to poor algorithm performance. 
Standard linear regression models are not suitable for this purpose whenever individual data points are not independent. Non-independence of the data is a typical characteristic of challenges, for example because multiple images from the same patient, or multiple frames from the same video, are used in the analysis. 
In such a situation, data are best represented by a hierarchical structure. In such a data tree (cf. Fig.~\ref{fig:hierarchy}), data corresponding to the same leaf can be assumed to be independent. Branches represent the source of non-independence, such as a specific hospital, device, or patient. To account for the correlation within challenge outcome data, we propose using Linear Mixed Models (LMMs) \cite{west2014linear}, as a generalization of linear regression models.

LMMs enable regressing an outcome using a linear combination of explanatory variables weighted by regression coefficients. The outcome variable is also referred to as \textit{dependent/target/response/explained variable} while the explanatory variables are also called \textit{independent variables}. In contrast to standard regression models, mixed models not only incorporate the parameters of interest, referred to as \textit{fixed effects}, but also so-called \textit{random effects} of variables explaining the hierarchical structure. 

In our specific (challenge) setting, image characteristics that have a potential influence on algorithm performance are by default assumed to be binary (present/not present; though categorical variables are also possible) and are represented as fixed effects in the model. Other factors, reflecting the hierarchical structure of the acquired data (e.g. the patient/hospital/image frame identifier) are represented as random effects. The fixed effects coefficients of the model then provide estimates of the impact of the provided image characteristics on the prediction performance. There are several ways to incorporate the challenge participant's algorithms in this setup. If the specific effects of the (typically small number of) algorithms are of interest, the algorithm can be modeled as a fixed effect, otherwise it is modeled as random effect. Alternatively, as a first approximation, aggregated metric values across algorithms (possibly after transformation; see below) may be used. 

Fig.~\ref{fig:statistical_modeling} presents a simplified workflow for choosing the appropriate mixed model for a given problem. The specific choice of the mixed model and the process of instantiating depends primarily on the distribution of the metric values. 
In the case of continuous metric values with unbounded support, an LMM \cite{west2014linear} is usually the most natural choice. 
Applied to the problem of challenge analysis, LMMs can be represented as:

\begin{equation}\label{eq:LMM}
    \overbrace{y}^{N \times 1} = \overbrace{\underbrace{\alpha}_{1 \times 1} \underbrace{\mathbf{1}}_{N \times 1}}^{N \times 1} + \overbrace{\underbrace{X}_{N \times p}\underbrace{\beta}_{p \times 1}}^{N \times 1} + \overbrace{\underbrace{Z}_{N \times q}\underbrace{u}_{q \times 1}}^{N \times 1} + \overbrace{\varepsilon}^{N \times 1}
\end{equation}
with the following components:

\paragraph{Outcome} The vector $y$ represents the $N$ metric values on the test images, which may be aggregated over all algorithms or be provided separately for all of them. For clarity of presentation, we will assume a single value per image in the following explanation, but an example of a more fine-grained composition is provided in Sec.~\ref{sec:methods:analysis_image_characteristics:quantification}.

\paragraph{Fixed effects} The so-called \textit{design matrix} $X$ corresponds to the $p$ image characteristics with corresponding $p$ fixed effects $\beta$. Generally, each row of $X$  represents an image and consists of $p$ binary variables representing the presence or absence of a specific image characteristic on the corresponding image. $\beta$ is the regression coefficient resulting from mixed model fitting, which can be used to predict the dependent variable $y$ from the fixed effects. Coefficients are easily interpretable in LMMs with untransformed dependent variables. Given, for example, presence of an image characteristic $c$, the expected metric value changes by $\beta_c$ compared to the situation when the characteristic is not present. The so-called \emph{intercept} $\alpha$  is a scalar representing an average outcome when all characteristics are absent (the arithmetic mean of $y$) and $\mathbf{1}$ is a vector of $1$s. 

\paragraph{Random effects} The matrix $Z$ is a so-called \textit{design matrix for the random effects}. Assuming, for example, a hierarchical data structure, in which clustering of test images arises from different patients, each patient would represent one of $q$ columns in $Z$, and each row in the matrix would have exactly one entry with 1 (0 otherwise), reflecting the fact that there is a unique assignment of outcome data to a specific patient. 
Similarly, further random effects, such as hospital IDs, can be incorporated by increasing the number of columns of $Z$ (and dimension of $u$). $u$ quantifies the random effect on the specific outcome. For example, the imaging device of a specific hospital could have a lower resolution, hence making predictions more difficult and thus leading to a worse metric value on average. Elements in $u$ are assumed to be normally distributed with zero mean and a variance interpreted as the between-cluster-variability (e.g. the variability in performance when comparing images of different hospitals). If the between-cluster-variability is large when compared to the residual variance (see below), observations within the cluster are highly correlated.

\paragraph{Residuals} $\varepsilon$ is a vector of residuals. The residuals are assumed to be normally distributed, $\mathcal{N}(0, \sigma_\varepsilon^2 I)$ capturing the variation in $y$ unexplained by the random and fixed effects.\\

Fitting of the LMM (i.e. estimation of the coefficients) is carried out through restricted maximum likelihood methods (REML). LMMs rely on the assumption of normality of the residuals, i.e., the outcome variable given the values of the explanatory variables follows a normal distribution. To detect a potential violation of the normality constraint, we recommend studying the Q-Q-Plot \cite{thode2002testing} of the residuals after model fitting.
Often, a transformation of the metric values must be applied to obtain approximate normality such that an LMM can be used. For a metric with values $\in$~$[0,\inf]$, for example, the \emph{log} transformation is a popular choice; if a metric is bounded by $[0,1]$, the \emph{logit} function, mapping the metric values to $[-\inf, \inf]$, is frequently used. Note that in case of such nonlinear transformation, additivity and linearity of effects on $y$ is lost, possibly complicating interpretation of regression coefficients.

If the outcome variable follows a non-normal distribution such as Binomial/Bernoulli, Poisson or Gamma, GLMMs, as a generalization of LMMs can be used (\cite{mcculloch2011generalized}). Specifically, if the outcome variable is binary, a mixed effects logistic regression model as a special case of GLMMs can be used.
An instantiation of such a model is provided in Sec.~\ref{sec:methods:analysis_image_characteristics:quantification}.

\subsection{Instantiation of the framework for multi-instance instrument segmentation}
\label{sec:methods:instantiation}
As a proof of concept, we instantiated our proposed framework for the ROBUST-MIS challenge 2019~\cite{ross2020robust}. This challenge was based on 10,040 endoscopic images that had been extracted from three different surgery types~\cite{maier2020heidelberg}. For each test case, the segmentation results of five (one algorithm was excluded due of a non-compliant training process that would affect comparability) participating algorithms were available.

\subsubsection{Annotation of image characteristics}
\label{sec:methods:analysis_image_characteristics:annotation}
To identify potential sources of error, we analyzed the general literature on endoscopic image analysis~\cite{maier2014comparative, ali2021bdeep} as well as the specific literature on artifacts in endoscopy~\cite{ali2020objective, ali2021adeep, funke2018generative, soberanis2020polyp} and on endoscopic vision challenges~\cite{bodenstedt2018comparative, allan20202018, ross2020robust}). Combined with personal experience gained during the annotation process for the challenge data~\cite{maier2020heidelberg}, we identified twelve relevant sources of error (see Table~\ref{tab:image_artifacts}) which were classified as global -- characterizing the whole image (here: dirty lens, overexposure and underexposure) -- or local (e.g. blood on specific instrument instance). The local features were provided for all instrument instances and/or the background individually.
A trained engineer with experience in endoscopic image annotation~(annotator was part of the annotation team of \cite{maier2020heidelberg}) then annotated the presence of such characteristics for all images (see Fig.~\ref{fig:image_artifacts_presence_in_datasets}). For the relevant test set (stage 3) this resulted in a total of $(5 + 3) \cdot 2,728 = 21,824$ image related and $9 \cdot 3,302 = 29,718$ instrument instance annotations.

\begin{table}[htb]
    \footnotesize
    \centering
    \caption{Meta data information provided by a human annotator for the entire background (Bg.) and for each instrument instance (Inst.), and/or globally.}
    \begin{tabular}{p{5cm}cc}
        \toprule
        \multicolumn{3}{c}{\textbf{LOCAL CHARACTERISTICS}}\\
        \textbf{\textit{Characteristic}}&  \textbf{\textit{Bg.}} & \textbf{\textit{Inst.}}\\
        \midrule
        Covered by blood? & \cmark & \cmark\\
        Covered by smoke? & \cmark & \cmark\\
        Covered by tissue? & \xmark & \cmark\\
        Subject to motion artifacts? & \cmark & \cmark\\
        Covered by specular reflections? & \cmark & \cmark\\
        Covered by another instrument? & \xmark & \cmark\\
        Covered by any other object (non-surgical)? & \cmark & \cmark\\
        Too bright? & \xmark & \cmark\\
        Too dark? & \xmark & \cmark\\
        \midrule
        \multicolumn{3}{c}{\textbf{GLOBAL CHARACTERISTICS}}\\
        \textbf{\textit{Characteristic}}&  \textbf{\textit{Img.}} &\\
        \midrule
        Is the image too bright? & \cmark&\\
        Is the image too dark? & \cmark&\\
        Does the lens seem dirty? & \cmark&\\
         \bottomrule
    \end{tabular}
    \label{tab:image_artifacts}
\end{table}

\subsubsection{Statistical analysis}
\label{sec:methods:analysis_image_characteristics:quantification}

The Dice Similarity Coefficient ($DSC$)~\cite{dice1945measures} is a widely used metric in medical image analysis~\cite{maier2018rankings, reinke2018exploit} and also served as a basis for the ranking in the ROBUST-MIS challenge~\cite{ross2020robust}. Yet, as $DSC$ values are in the range of $[0, 1]$, modeling the algorithm performances directly on the challenge metric would violate the normality assumption of the residuals $\varepsilon$. 
As mentioned in Sec.~\ref{sssec:MM-analysis}, the problem can potentially be overcome by applying a transformation $f(\cdot)$, such as the \emph{logit} to metric values bounded in  $[0, 1]$, with the goal of mapping the values to the range of $[-\inf, \inf]$. 
As this process did not yield an approximate normal residual distribution of $\varepsilon$ 
for our data, we propose regarding the segmentation problem as a pixel-level classification problem and applying a GLMM to model the target metrics precision and recall as a function of image characteristics, as detailed in the following paragraphs.

\paragraph{Outcome reformulation} For this study, we leverage the fact, that the $DSC$ is closely related to \textit{precision} ($PRE$) and \textit{recall} ($REC$). More specifically, $DSC$, $PRE$ and $REC$ can all be expressed as a function of the number of true(T)/false(F) positives(P)/negatives(N):

\begin{equation}
    PRE =  \frac{TP}{TP + FP}; ~ REC = \frac{TP}{TP + FN}
\end{equation}

\begin{equation}
    DSC = F1 = \frac{2 TP}{2 TP + FP + FN} = \frac{2 \cdot PRE \cdot REC}{PRE + REC}
\end{equation}

To use \emph{precision} and \emph{recall} as target metrics, we convert each multi-instance reference annotation mask into a set of binary masks, each corresponding to one instrument instance. For each image $i \in I$, each instrument $j \in J_i$ that is present in $i$ and each algorithm $k \in K$ we then determine both, the \textit{recall}, defined as the probability $\pi_{i, j, k}$ of a pixel of the reference segmentation to be present in the mask provided by the algorithm, and the \emph{precision}, defined as the probability $\tilde{\pi}_{i, j, k}$ of a pixel of the segmentation mask to be present in the reference segmentation. In other words, we relate the TP to either the reference mask of an instance (\textit{recall}) or the mask provided by the algorithm (\textit{precision}). Formally, the pixel-level classification (per instance) is binary and thus can be regarded as a Bernoulli experiment. Depending on the perspective (\textit{recall} or \textit{precision}) if a pixel is correctly classified as instrument follows a Bernoulli distribution $Bernoulli(\pi)$ with parameter $\pi = \pi_{i, j, k}$ resp. $\pi = \tilde{\pi}_{i, j, k}$. 

\paragraph{GLMM fitting} 
In GLMMs, a \emph{link function} $g$ relates the expected outcome (here the parameter $\pi$) with the linear predictor. The canonical choice for this function in case of binary data is the logit link function $g(\pi):=\log\frac{\pi}{1-\pi}$. The complete equation is then given by

\begin{equation}\label{eq:GLMM}
\overbrace{g(\pmb{\pi})}^{N \times 1} = \overbrace{\underbrace{\alpha}_{1 \times 1} \underbrace{\mathbf{1}}_{N \times 1}}^{N \times 1} + \overbrace{\underbrace{X}_{N \times p}\underbrace{\beta}_{p \times 1}}^{N \times 1} + \overbrace{\underbrace{Z}_{N \times q}\underbrace{u}_{q \times 1}}^{N \times 1},
\end{equation}

where $\pmb{\pi}$ is a column-vector consisting of all the probabilities $\pi_{i, j, k}$ (resp. $\tilde{\pi}_{i, j, k}$) and $g$ is applied element-wise. As described above, we define fixed effects $\beta$ for image and instrument characteristics (within $X$), resulting in $p = 17$. Furthermore, we model the participants' algorithms, as well as the patient, image and instrument as independent random effects combined in vector $u$ (with indicator matrices combined in $Z$), resulting in $q = |K| + |P| + |I| + \sum_{i \in I} |J_i|$, where $|K|, |P|, |I|, \sum_{i \in I} |J_i|$ refer to the number of algorithms, the number of patients, the number of images and the total number of instances, respectively. The number of rows $N$ in our model comes down to $N = |K| \cdot \sum_{i \in I}|J_i|$. Note that we inspected the cases of single and multiple instruments per image separately (see Tab.~\ref{tab:image_effect_characteristics}). So while $|K| = 5$ and $|P| = 10$ stayed constant there were $1,184$ images with $1$ instrument and $1,031$ images with multiple instruments each ($2,118$ instruments in total).

\paragraph{Model interpretation} After fitting the model, coefficients $\beta$ can be interpreted in terms of log odds ratios (OR)~\cite{mcculloch2011generalized}. The OR is a statistic measuring how two events are associated with each other regarding their presence or absence~\cite{mcculloch2011generalized}. Here, the OR measures the ratio of the odds (e.g. $\pi_{i, j, k} / (1-\pi_{i, j, k})$) in the presence and absence of a given image characteristic. 
However, the fact that the OR is not symmetrical around 1 (the value indicating no effect of the image characteristic) makes the comparison of the individual effects less intuitive (as can be illustrated by an OR of 2:3 ($0.\overline{6}$) and its inverted ratio of 3:2 ($1.5$)). 
Fig.~\ref{fig:image_artifacts:more_than_two_instances} thus shows the $\log$ of the OR, making the values symmetrical around 0. For the interpretation, a positive log OR increases the chance that a high metric is measured, while a negative log OR decreases the chance.

\subsection{Strength-weakness-driven algorithm development}
\label{sec:methods:algorithm_development}

\begin{figure*}[tbp]
    \centering
    \includegraphics[trim=80 150 200 200, clip, width=1\textwidth]{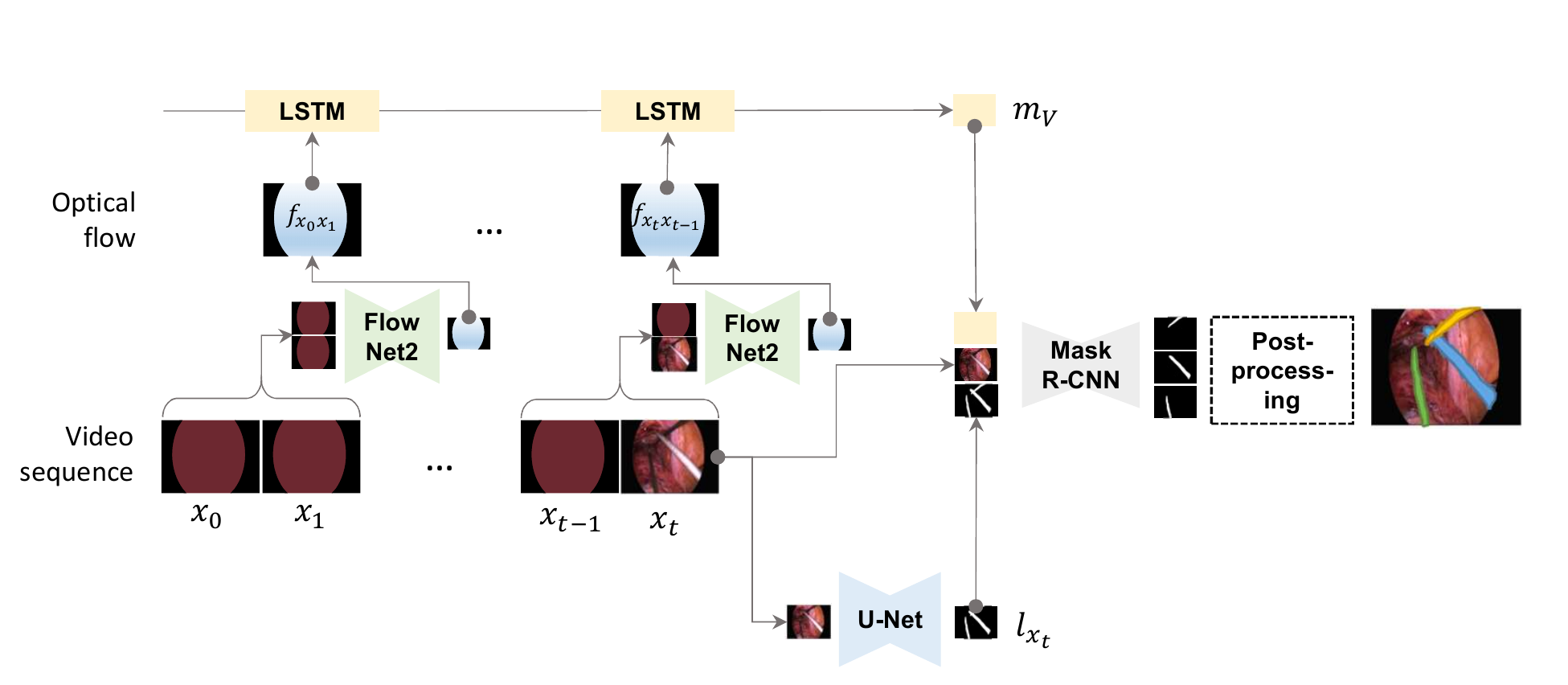}
\caption{
    Concept of the multi-instance segmentation approach for a video frame $x_t$ of a video $V$. A Mask R-CNN performs instance localization based on three input channels: The current image frame $x_t$, the instrument likelihood $l_{x_t}$ estimated with a U-Net, and third channel $m_V$ encoding motion information via optical flow. The optical flow of a video sequence $V=(x_{0}, \dots, x_t)$ of length $t + 1$ is estimated by (1) determining the optical flow for each pair of consecutive frames via FlowNet2~\cite{ilg_flownet_2017} and then (2) summarizing the pairwise information with an LSTM into $m_V$. A post-processing step applied to the output of the Mask R-CNN (see section \ref{subsec:post_processing}) yields the final result.
    }
    \label{fig:architecture_overview}
\end{figure*}

As detailed in the results section (Sec.~\ref{sec:artifact_analysis:e1_influence_of_artifacts}), the GLMM revealed several image characteristics with major impact on algorithm performance, namely: motion and underexposure of the instrument, crossing medical instruments as well as smoke or other medical equipment (e.g. swabs or bandages) in the field of view. We hypothesized that a majority of these issues can be addressed by going beyond single image analysis and \textit{taking the temporal context into account}. In fact, the annotators of the Heidelberg Colorectal Data Set for Surgical Data Science in the Sensor Operating Room (HeiCo)~\cite{maier2020heidelberg}, which served as basis for the ROBUST-MIS challenge, also reported that analysis of preceding frames was sometimes necessary to label a given frame. The proposed architecture resulting from the GLMM analysis is shown in Fig.~\ref{fig:architecture_overview}. The core component of the presented deep learning architecture is a masked region-based convolutional neural network (Mask R-CNN)~\cite{he2017mask} that uses the following information as input: (1) the raw video frame, (2) the probability of a pixel to be an instrument and (3) the Long short-term memory (LSTM)-summarized information on object motion encoded in an optical flow. Details are provided in \ref{app:sec:algorithm_development}.

\section{Experiments and Results}
\label{sec:experiments}

We validated our framework on the ROBUST-MIS challenge 2019 data set~\cite{ross2020robust, maier2020heidelberg}. The following sections present our findings with respect to image characteristics impacting algorithm performance as well as the validation of our algorithm tailored to the specific weaknesses of the state of the art.

\subsection{Effect of image characteristics on the performance of state-of-the-art algorithms}
\label{sec:artifact_analysis:e1_influence_of_artifacts}

\begin{figure}[htb]
    \centering
    \includegraphics[trim=75 71 75 100, clip, width=0.8\linewidth]{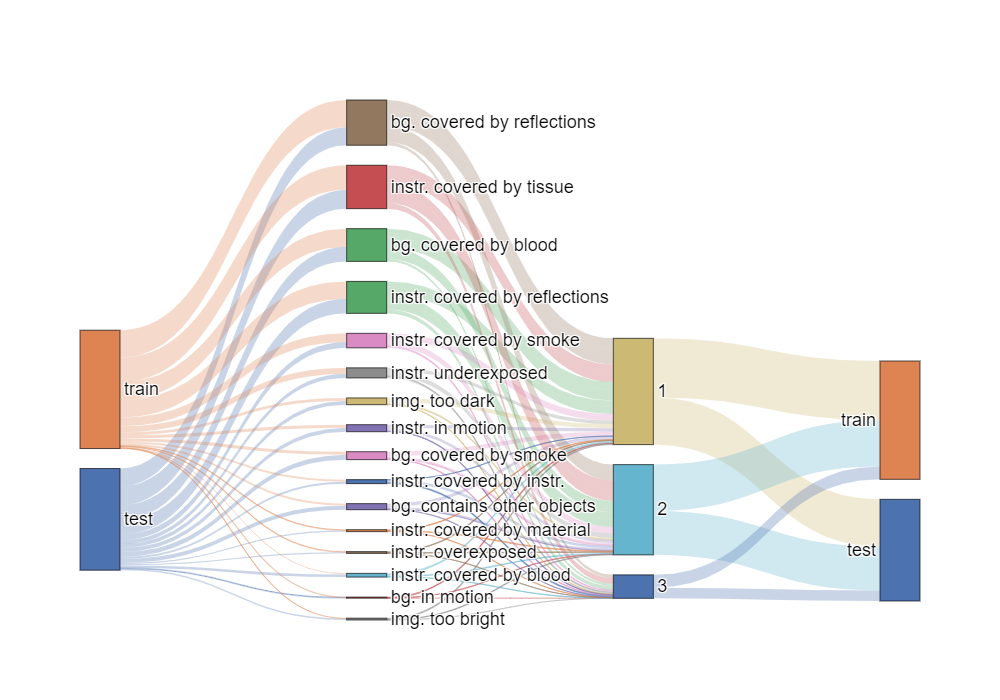}
    \caption[Presence of image characteristics in training and test data]{Summary of the 195,148 meta data annotations performed on the ROBUST-MIS challenge data (training and test data). The frequency of the selected image characteristics (Tab.~\ref{tab:image_artifacts}) in the test and training data set is shown along with information on the number of instruments per image. The size of the blocks and connecting lines in the image correspond to the proportion of images that have the respective property. The colors were picked to improve the legibility of the image. ("bg.": background; "instr.": instrument; "img.": image)}
    \label{fig:image_artifacts_presence_in_datasets}
\end{figure}

The frequency of special image characteristics captured by the meta data annotation is shown in Fig.~\ref{fig:image_artifacts_presence_in_datasets}.

\begin{figure*}[htb]
    \centering
    \includegraphics[trim=0 0 0 0, clip, width=0.8\linewidth]{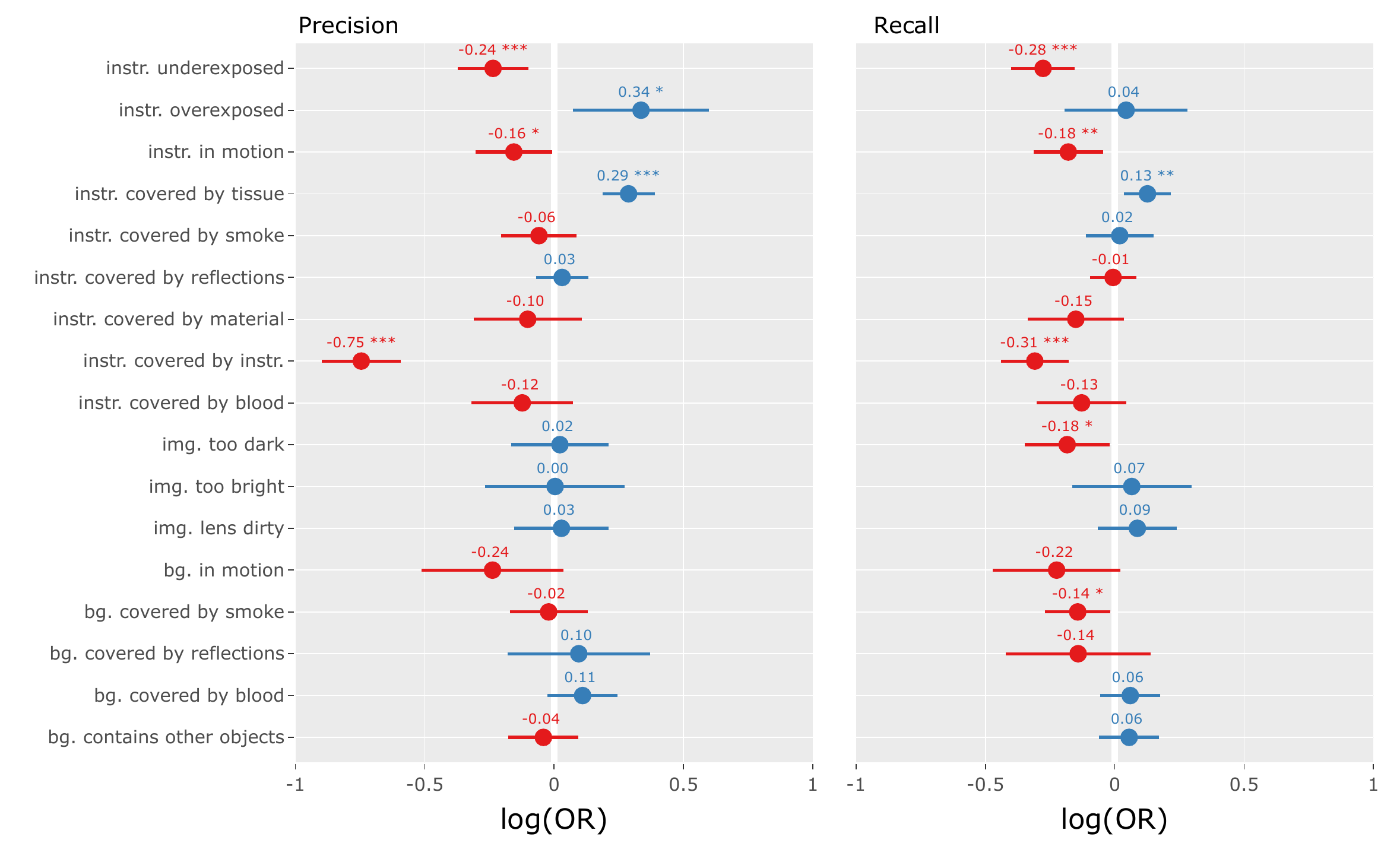}
    \caption{Impact of image characteristics on the algorithm performance for $>=2$ instances. The characteristics' effect is displayed in the form of the $\log(OR)$, representing the logarithms of the odds of occurrence of the outcome in presence of the image characteristic, compared to the odds of the outcome occurring in the absence of that exposure. Significant effects are marked with an asterisk ($\ast: p<=0.05$, $\ast\ast: p<=0.01$, $\ast\ast\ast: p<=0.001$).} \label{fig:image_artifacts:more_than_two_instances}
\end{figure*}

Following the statistical methodology presented in Sec.~\ref{sec:methods:instantiation}, we analyzed the influence of image characteristics on the algorithm performance, using precision and recall as metrics. The main results are presented in Fig.~\ref{fig:image_artifacts:more_than_two_instances} and Tab.~\ref{tab:image_effect_characteristics}.



\begin{figure*}
    \centering
    \includegraphics[width=\textwidth]{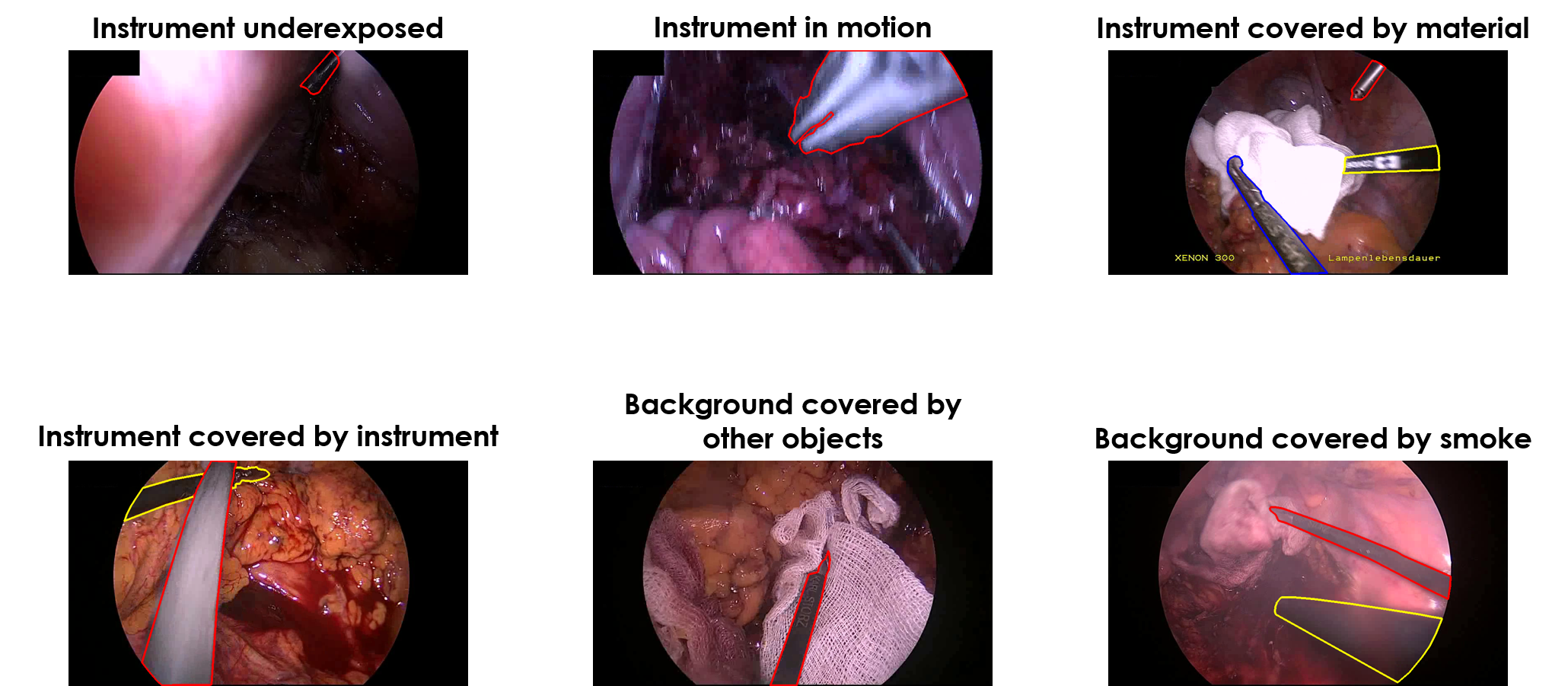}
    \caption{Examples for identified main image characteristics that impact the algorithm performance.}
    \label{fig:image_effect_examples}
\end{figure*}

According to our results, the following main characteristics had a statistically significantly ($p<0.05$) negative influence on the results: instrument is underexposed, in motion or covered by material, or background is covered by smoke or other objects. Example frames are provided in Fig.~\ref{fig:image_effect_examples}. 
When two or more instrument instances were visible (see Fig.~\ref{fig:image_artifacts:more_than_two_instances}), the statistically significant characteristic with 
the largest impact was "instrument covered by another instrument". 

\begin{table*}
    \scriptsize
    \centering
    \caption{Sources of algorithm failures and successes, where $+$ denotes a significant positive effect of $0<OR\leq 1.25$, $++$ a significant positive effect of $1.25<OR\leq 1.50$ and $+++$ a significant positive effect of $1.50<OR$. Analogously, $-$ denotes a significant negative effect of $0.75<= OR < 0$, $--$ a significant negative effect of $0.50 \leq OR < 0.75 $ and $--$ a significant negative effect of $OR< 0.50$. Empty columns indicate no significant impact could be found.  A significance level of 0.05 is used throughout.  "x" means that the effect could not be assessed (e.g., the effect covered by instrument does only exist when $n>1$).}
    \begin{tabular}{p{3.35cm}|cc|cc}
        \toprule
        & \multicolumn{4}{c}{\textbf{\textit{LOCAL CHARACTERISTICS}}} \\
        &  \multicolumn{2}{c|}{\textbf{\textit{PRECISION}}} & 
        \multicolumn{2}{c}{\textbf{\textit{RECALL}}}\\
        \textbf{\textit{Characteristic}} & inst. > 1 & inst. = 1 & inst. > 1 & inst. = 1\\
        \midrule
        instrument overexposed & $++$ & & &\\
        instrument covered by tissue & $++$ & & $+$ & $++$\\
        instrument underexposed & $-$ & $--$ & $--$ & $---$ \\
        instrument covered by reflections & & $++$ & & \\
        instrument covered by material & & $--$ & & $--$ \\
        instrument covered by smoke & & & & \\
        instrument in motion&  & $--$ & $-$ &  \\
        instrument covered by blood& & & & \\
        instrument covered by instrument & $---$ & x & $--$ & x\\
        background contains other objects & & $--$ & & \\
        background in motion & & & & \\
        background covered by blood & & & & \\
        background covered by reflections & & & & \\
        background covered by smoke & & $-$&  $-$ & \\
        \midrule
        & \multicolumn{4}{c}{\textbf{\textit{GLOBAL CHARACTERISTICS}}} \\
        &  \multicolumn{2}{c|}{\textbf{\textit{PRECISION}}} & 
        \multicolumn{2}{c}{\textbf{\textit{RECALL}}}\\
        \textbf{\textit{Characteristic}} & inst. > 1 & inst. = 1 & inst. > 1 & inst. = 1 \\
        \midrule
        image lens dirty & & & & \\
        image too bright & & & & \\
        image too dark & & &$-$ &\\
         \bottomrule
    \end{tabular}
    \label{tab:image_effect_characteristics}
\end{table*}

\subsection{Performance of algorithm developed based on strength-weakness-analysis}

\begin{figure}[htb]
    \centering
    \includegraphics[trim=0 50 0 25, clip,width=1\linewidth]{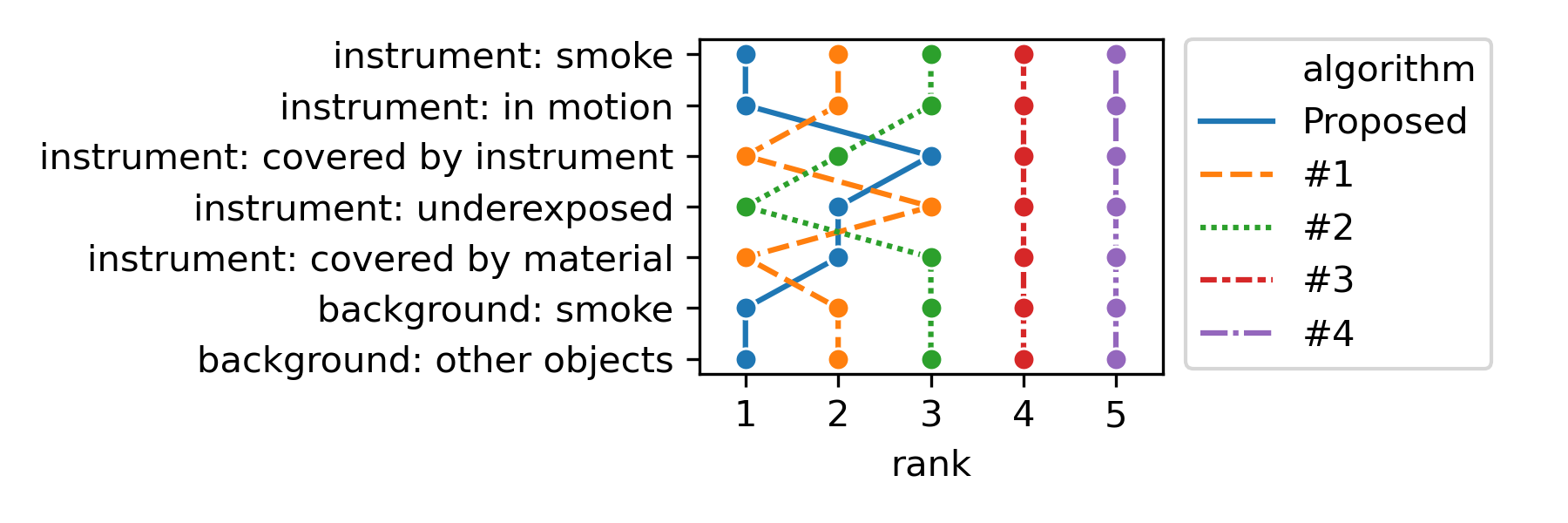}

    \caption{Comparison of our method ($T^{+}_{RFL}$) with the four best-performing methods of the ROBUST-MIS challenge (\#1-\#4)  for the seven characteristics that have a significant negative impact on algorithm performance according to our effect analysis. It can be seen that our proposed method ($T^+_{RFL}$) outperforms the competitors in most categories.}
    \label{fig:characteristic_driven_development:characteristic_ranking}
\end{figure}

Our method resulting from the strength-weakness-analysis was compared to the ROBUST-MIS 2019 challenge's top-scoring methods based on the accuracy as reported in the ROBUST-MIS challenge~\cite{ross2020robust} and using the challengeR analysis tool~\cite{wiesenfarth2021methods}. For the validation of our method, we used the metrics proposed by the ROBUST-MIS challenge as this allowed us to compare it to the best-performing related methods. Following the challenge guidelines \cite{ross2020robust}, we split the  data into 5,983 training and 4,057 test images. The reason for the relatively high number of test images compared to training images was the fact that we reserved one surgery type exclusively for testing, as detailed in~\cite{ross2020robust}. As shown in Fig.~\ref{fig:characteristic_driven_development:characteristic_ranking}, our method outperformed the state of the art in the majority of categories that represent typical failure cases. 

\begin{figure}[htb]
    \centering
    \includegraphics[width=0.48\textwidth]{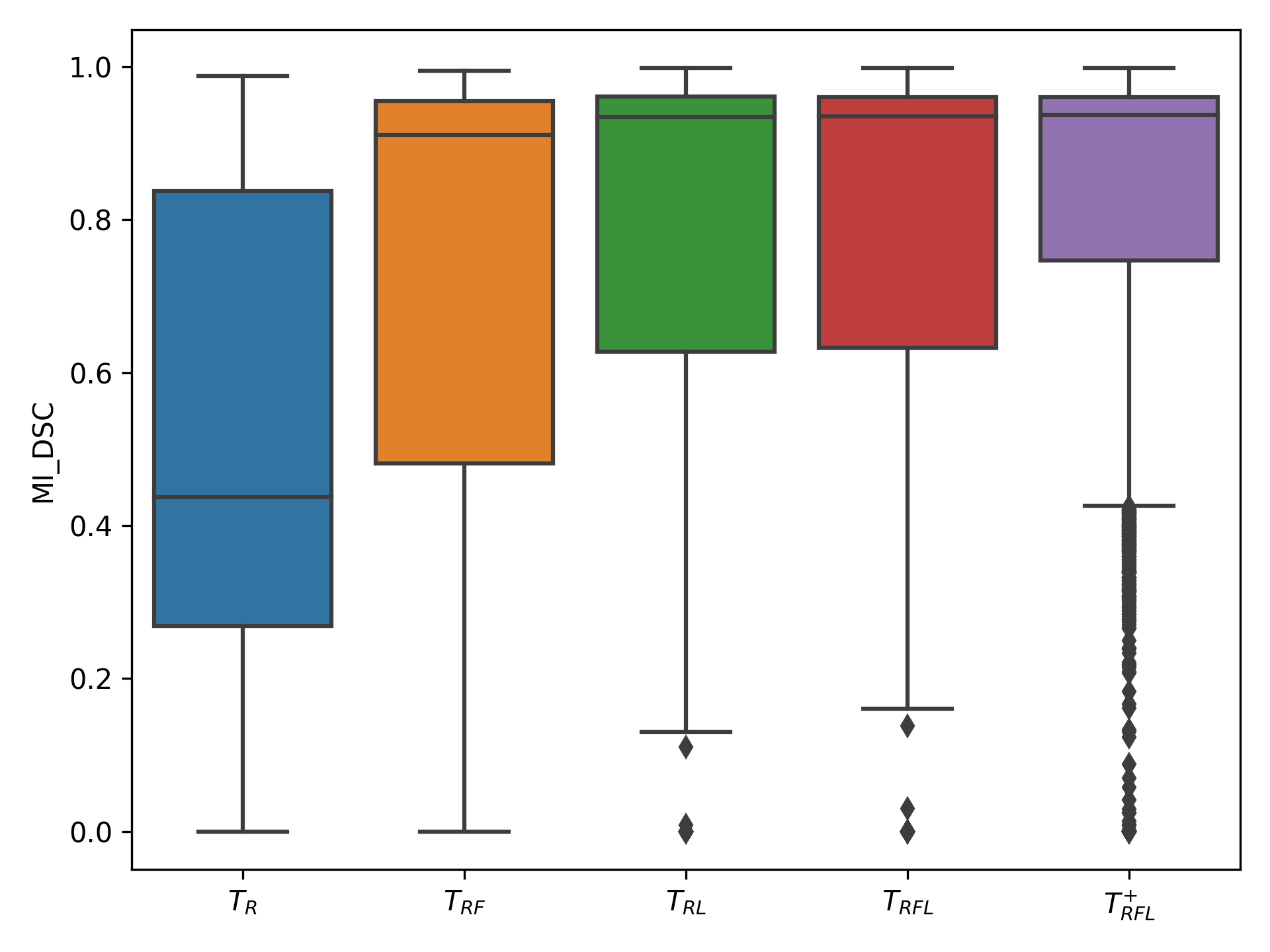}
    \caption[Effect of including temporal information and instrument likelihood]{Performance of the model for different input data. $T_R$: only (raw) input images; $T_{RF}$ images and flow; $T_{RL}$ images and instrument likelihood; $T_{RFL}$ images, flow and likelihood. $T_{RFL}^{+}$ indicates using an additional post-processing step after applying ($T_{RFL}$).}
    \label{fig:characteristic_driven_development:ablation_study}
\end{figure}

We further performed an ablation study to assess the benefit of the different architectures components, namely  (1) the optical flow and (2) the instrument likelihood as additional input to the Mask R-CNN and (3) our post-processing method applied to the output of the Mask R-CNN. The results are presented in Fig.~\ref{fig:characteristic_driven_development:ablation_study}. Including the optical flow ($T_{R}$ vs. $T_{RF}$) improved the performance by a factor of $2$, from a median \textit{MI\_DSC} of  $0.44$ to $0.91$ (mean: $0.50$ to $0.73$ ($146\%$)). Including the instrument likelihood in the network model ($T_{RL}$) also increased the median \textit{MI\_DSC} by a factor of $2$ ($0.44$ to $0.93$ ($211\%$); mean $0.50$ to $0.79$ ($158\%$)). Incorporating both the flow and the instrument likelihood as additional input ($T_{RFL}$) did not yield a substantial improvement compared to $T_{RL}$.
The post-processing significantly ($p=2.7E-7$) increases the mean  performance of the model $T_{RFL}$ at $1\%$, while simultaneously reducing the IQR from $0.33$ to $0.21$. Also, the robustness of the method (defined in~\cite{ross2020robust} as the 5th percentile) increased from $0.28$ to $0.32$. Further descriptive statistics can be found in Tab.~\ref{tab:characteristic_driven_development:ablation_study}.\\

\begin{table}
    \centering
    \caption{Effect of the different inputs for training the Mask R-CNN, showing the mean ($\mu$), median ($\widetilde{x}$), 5th, 25th, and 75th quartile ($Q_1$, $Q_3$), and the interquartile range ($IQR$) of the multi-instance dice coefficient $DSC_{MI}$. The names of the trained model $T$ with the indices $R$, $L$ and $F$ are referring to $R=\text{raw}$, $F=\text{flow}$ and $L=\text{likelihood}$, as detailed in the App.~\ref{app:sec:algorithm_development}. The model $T_{RFL}^{+}$ is the same model as $T_{RFL}$, but followed by the post-processing.}
    \begin{tabular}{l|ccccccc}
         \toprule
         \textbf{Model} &
         \textbf{$\mu$} &
         \textbf{$\widetilde{x}$} &
         \textbf{$Q_{5}$}&
         \textbf{$Q_{25}$}&
         \textbf{$Q_{75}$}&
         \textbf{$IQR$}
         \\
         \midrule
         $T_R$&
         0.50&
         0.44&
         0.00&
         0.27&
         0.83&
         0.56&
         \\
         $T_{RF}$&
         0.73&
         0.91&
         0.0&
         0.48&
         0.95&
         0.47
         \\
         $T_{RL}$&
         0.79&
         0.93&
         0.24&
         0.63&
         0.96&
         0.33
         \\
         $T_{FL}$&
         0.80&
         0.93&
         0.29&
         0.63&
         0.96&
         0.33
         \\
         $T_{RFL}$&
         0.80&
         0.93&
         0.29&
         0.63&
         0.96&
         0.33
         \\
         $T_{RFL}^+$&
         \textit{0.81}&
         \textit{0.94}&
         \textit{0.32}&
         \textit{0.75}&
         \textit{0.96}&
         \textit{0.21}\\
         \bottomrule
    \end{tabular}
    \label{tab:characteristic_driven_development:ablation_study}
\end{table}

\section{Discussion}
\label{sec:discussion}
While a lot of research is currently invested in maximizing algorithm performance for various image analysis tasks, comparatively little effort is currently put into failure analysis. This holds especially true for the growing field of benchmarking via challenges. Although challenge results potentially encode crucial information with respect to typical failure cases as well as reasons for algorithm failure, most challenge organizers restrict their reports to plain ranking tables~\cite{wiesenfarth2021methods} leaving the rich challenge data unexploited. To address this gap in the literature, we presented a statistical framework for systematically learning from challenge results and instantiated it for the specific task of medical instrument segmentation. The following sections discuss the general approach (Sec. \ref{sec:discussion:framework}), the specific findings for the sample application (Sec. \ref{sec:discussion:instanstiation_robust_mis}) as well the new algorithm resulting from the statistical analysis (Sec. \ref{sec:discussion:algorithm_developed}).

\subsection{Framework for challenge analysis}
\label{sec:discussion:framework}

In the field of biomedical image processing, failure analysis is often restricted to qualitative assessments~\cite{wiesenfarth2021methods}. Even quantitative analyses are typically restricted to single parameter assessments contrasting algorithm performance in the presence or absence of certain features (e.g.~\cite{soberanis2020polyp}). This approach should be used cautiously as it neglects a possible strong interplay between features, thus potentially confusing correlation with causation. We overcome these limitations by a statistically well-grounded approach. Leveraging the strengths of mixed models, we are able to disentangle the effects of different sources and clearly identify sources of algorithm failure. This enables the design of future algorithms dedicated to the actual needs. 

A limitation of our concept is the additional manual annotation effort involved. Depending on the specific application, this could potentially be overcome by an automatic annotation of image characteristics, or by (quality-controlled) crowdsourcing~\cite{maier2014crowdsourcing, heim2018large}. It should further be noted that we instantiated the concept only for a single challenge, with specific focus on multi-instance medical instrument segmentation. However, as segmentation is by far the most widely used task in challenges~\cite{maier2018rankings}, our concept can be transferred to a majority of studies and also create awareness of the problem that arises when using statistical tests for bounded metrics. Finally, it must be noted that statistical modelling is a complex process of balancing adherence to specific assumptions and model complexity. Delivering a recipe applicable to every challenge is not possible within the scope of a single paper but we believe that this work could trigger more sophisticated challenge analyses following the general approach presented here.

\subsection{Instantiation for ROBUST-MIS challenge}
\label{sec:discussion:instanstiation_robust_mis}
In contrast to the literature (e.g., \cite{maier2014comparative, ali2021bdeep, ali2020objective, ali2021adeep, funke2018generative, soberanis2020polyp,bodenstedt2018comparative, allan20202018, ross2020robust}), our study did not reveal a harmful effect of reflections, blood, and smoke on the algorithm performances. Instead, our analysis showed that the main limiting factors are when an instrument is in motion, underexposed, or covered by another instrument. Interestingly, other characteristics such as an instrument being overexposed or covered by tissue, reflections or blood, seem to even support the algorithm performances. That artifacts/characteristics are not always harmful is in line with a recent work from \cite{kayser2020understanding} who use reflections to improve the segmentation of polyps. 
The fact that the characteristic "instrument covered by tissue" has a slight positive effect on algorithm performence is most likely due to the fact that the visible tissue overlay mainly occurs when the instrument is clearly visible and distinguishable from the background. 

While most of the characteristics either harm or benefit both metrics investigated, we identified one characteristic that yielded different results for precision and recall. Specifically, when the background contains other objects, we observe an increased recall but a decreased precision. This indicates that an oversegmentation occurred typically occurred in these cases. 


The strongest negative impact by far was found when the instrument is being covered by another instrument. This can be attributed to the architecture of Mask R-CNNs and their way of processing images. A Mask R-CNN relies on a region proposal network that provides bounding boxes around regions of particular interest. Especially in regions where instruments overlap, those bounding boxes might contain parts of multiple instrument instances, which then leads to poor segmentations. This finding is in line with work on the Mask R-CNN problem of overlapping instances~\cite{xu2020maskplus}.


\subsection{Algorithm development tailored to failure cases}
\label{sec:discussion:algorithm_developed}
With a few exceptions~\cite{hasan2019u, isensee2020or}), the few methods published on multi-instance segmentation to date use a Mask R-CNN \cite{gonzalez2020isinet, kletz2019identifying} as core component. To our knowledge, Kletz et al.~\cite{kletz2019identifying} were the first to use a Mask R-CNN for surgical instrument segmentation. They developed their method for laparoscopic gynecology videos but reported limitations under conditions of occlusion and overlapping instruments. In the context of binary segmentation, several authors have aimed to tackle this issue by including temporal information in order to improve their segmentation, e.g., in situations where instruments are only partially visible, due to overlapping tissue. Jin et al.~\cite{jin2019incorporating} were the first who estimated the optical flow by using a CNN, namely UnFlow~\cite{meister2018unflow}, for the segmentation of instruments. Instead of using the optical flow as an additional feature, they used it as a prior for initializing the attention of a temporal attention pyramid network to learn to focus on moving objects. However, their approach was for binary segmentation, type classification and instrument parts segmentation and classification, thus requiring a much simpler network architecture and no management of pixels that could possibly belong to different object instances.

The first to combine a Mask R-CNN and the optical flow were  Gonz{\'a}lez et al.~\cite{gonzalez2020isinet}, who proposed the use of a Mask-RCNN  to ensure a segmentation and classification of instruments as, e.g., grasper or scissors. The authors computed the optical flow of previous frames to include a temporal consistency module and consider an instance's predictions across the frames in a sequence. The approach outperformed the state-of-the-art methods on the Endoscopic Vision 2017 and 2018 Robotic Instrument Segmentation data sets~\cite{allan20192017}. However, this work was also not used for multi-instance segmentation.

Up to this point, and to the best of our knowledge, none of the approaches for multi-instance segmentation have successfully incorporated temporal information in the algorithm, as further reported by~\cite{ross2020robust}.
To address this gap in the literature, a new algorithm was developed that generated a benefit using optical flow in combination with instrument probability in order to explicitly address the previously mentioned weaknesses. While we found a huge performance boost when integrating the flow as an additional input of a native Mask R-CNN, the effect was negligible when also incorporating instrument likelikehood. In contrast, we achieved a high performance boost with an additional post-processing step dedicated to resolving ambiguities in the presence of overlapping instruments.

The presented results suggest that typical challenges of laparoscopic videos, such as reflection, blood, and lighting variations, are already well manageable by state-of-the-art methods. However, difficulties with tube-like structures that are misclassified as instruments or transparent objects such as trocars persist. Furthermore, images with crossing or close instruments remain difficult to separate for both  state-of-the-art methods and the presented approach, even though the latter was specially designed to manage such difficulties. One limiting factor may be seen in restrictions in the training and test data set, where only 8\% of the images contain more than two instrument instances. Furthermore, only in rare cases do those instances overlap or intersect, thus resulting in only limited opportunities for training and evaluating the algorithm's separation capabilities.

It should be mentioned that real-time capability is an essential prerequisite for successful translation to a clinical setting. Currently, the Conditional Random Field (CRF) \ref{subsec:post_processing} and the estimation of the optical flow would already approximately take more than 2 seconds per image. However, the method presented was merely an attempt to solve the multi-instance segmentation problem. The next step could be rendering the algorithm real-time capable.

Overall, our methodology achieved a new best score on the ROBUST-MIS challenge data set. While we did not use the challenge test data to tune hyperparameters, it should be mentioned that we had access to the other participants' performance results to inform the strength-weakness driven algorithm development. This could still be seen as a competitive advantage. However, the primary aim of this study was not to present a new state-of-the-art method for instrument segmentation but a novel concept for learning from challenges. With this work, we have not only identified typical failure cases for the task of medical instrument segmentation but also showcased an entirely new way of problem-driven algorithm development based on insights gained through challenge results.

\section{Conclusion}
\label{sec:conclusion}

In conclusion, the proposed approach to leveraging meta data annotations for a mixed model-based analysis of challenge results opens up entirely new opportunities for systematically learning from challenges. By identification of characteristics that lead to algorithm failure it not only provides a much deeper understanding of the state of the art for a given application but also enables tailoring future algorithm development to the actual remaining needs.


\section*{Acknowledgments and conflicts of interest}
The authors would like to thank Minu Dietlinde Tizabi and Alexander Seitel for proofreading the paper. Part of this work was funded by the Helmholtz Imaging Platform (HIP), a platform of the Helmholtz Incubator on Information and Data Science and by the Surgical Oncology Program of the National Center for Tumor Diseases (NCT) Heidelberg. Finally, thanks to all authors of the ROBUST-MIS challenge manuscript~\cite{robustmisgrand-chall}, which served as foundation of this work.

\bibliographystyle{unsrt}
\bibliography{references}

\section*{Appendix}
\appendix

\section{Segmentation algorithm}
\label{app:sec:algorithm_development}
Formally, the multi-instance instrument segmentation problem is defined as follows:
Given is a video sequence $V=(x_{0}, x_{1}, \dots, x_t)$ that consists of $t + 1$ frames. All frames have the dimensions $[h \times w \times 3]$, where $h$ and $w$ are the image height and width, and the three channels in the last dimension contain the RGB-encoded color information.
To the final frame $x_t$ of $V$ corresponds a set of instrument instances $J_{V}$ present in frame $x_t$, that have to be segmented.

Our basic approach was briefly presented in Sec.~\ref{sec:methods:algorithm_development} and illustrated in our overview Fig.~\ref{fig:architecture_overview}. 
The following sections present the details of the instrument likelihood prediction in Sec.~\ref{subsec:prediction_instrument_likelihood}, the method for flow estimation in Sec.~\ref{subsec:prediction_optical_flow}, the final prediction of instrument instances in Sec.~\ref{subsec:multiple_instance_segmentation} and the post-processing in Sec.~\ref{subsec:post_processing}.

\subsection{Instrument likelihood prediction}
\label{subsec:prediction_instrument_likelihood}
Given a video sequence $V$, the goal of this step is to predict within the final frame $x_t$ of $V$ for each pixel in $x_t$ the probability that it is an instrument. Arranged as the original image shape $[h \times w \times 1]$ this yields the likelihood map $l_{x_t} := \mathcal{B}(x_t)$ as output of the binary segmentation model $\mathcal{B}$. Prediction is done with a 2D U-Net~\cite{ronneberger2015u}, which achieved the best results in the ROBUST-MIS binary segmentation challenge~\cite{ross2020robust} and is further described in Isensee et al.~\cite{isensee2020or}.\\

\subsection{Usage of the temporal information}
For segmenting instrument instances in the final frame $x_{t}$ of a video sequence $V$, the goal is to include the knowledge of the previous frames $x_{0}, ..., x_{t-1}$. In the first step, the instrument motion is estimated using the optical flow concept. In the second step, this movement is summarized so as to enable straightforward processing in the segmentation task. The following two paragraphs will explain how the optical flow is generated and summarized for later use in the segmentation network.

\paragraph{Prediction of the optical flow}
\label{subsec:prediction_optical_flow}
Prediction of the optical flow $f_{x_{i+1}, x_{i}}$ is done for two consecutive frames $x_i$ and $x_{i+1}$ at a given time step $0 \leq i < t$ via FlowNet2~\cite{ilg_flownet_2017} $\mathcal{F}$, such that $\mathcal{F}(x_{i+1}, x_{i}) = f_{x_{i+1}, x_{i}}$. The dimensions of $f_{x_{i+1}, x_{i}}$ are $[h \times w \times 2]$. The two layers in the last dimension correspond to the shift of each pixel in x- (first layer) and y- (second layer) direction. Estimation of the optical flow is defined to be backwards in time, from $x_{i+1}$, to $x_{i}$. Defining this is important because the optical flow estimation is not symmetric due to effects such as occluded regions~\cite{sanchez2015computing}.

\paragraph{Summarizing the optical flow}

To summarize the optical flow $m_V$ as a latent representation of the motion in the complete sequence $V$ of length $t+1$, as model $\mathcal{R}(V, t)$, a recurrent neural network in the form of a LSTM was chosen~\cite{hochreiter1997long}. We decided to summarize the optical flow separately, but with the same network, in the x- and y- direction, which led to the latent representation $m_V$ with the dimensions $[h \times w \times 2]$:

\begin{equation}
    \begin{aligned}
        m_V = \mathcal{R}((\mathcal{F}(x_{i+1}, x_{i}))_{0 \leq i < t }), t).
        \label{eq:summarizing_optical_flow}
    \end{aligned}
\end{equation}

\subsection{Segmentation of multiple instances}\label{subsec:multiple_instance_segmentation}
Recall, we are given a video sequence $V$ of length $t+1$, where for the last video frame $x_{t}$, all instrument instances $J$ that are present in $x_t$ should be segmented. As the number of visible instances differs from image to image, a Mask R-CNN was used as a segmentation model $\mathcal{S}$. To keep the same spatial dimensions, the input of $\mathcal{S}$ is the concatenation in the last dimension of the RGB image $x_t$, the likelihood of a pixel being an instrument $l_{x_t}$ and the latent space of the motion over all video frames $m_V$:
\begin{equation}
    J_{V} = \mathcal{S}(x_t, l_{x_t}, m_V),
    \label{eq:mi_segmentation}
\end{equation}
where $J_{V}=\left\{(j_{V,i}, s_{V,i})| 1 \leq i \leq N_{V}\} \right\}$ is a set of predictions that yields pairs of predicted instances $j_{V,i}$ with its corresponding predicted Intersection over Union ($IoU$) score~\cite{jaccard1912distribution} $s_{V,j}$ and $N_{V}$ being the number of predicted instances. The dimensions of a $j_{V,i}$ are $[h \times w\times 1]$, with values within $[0, 1]$ that are the probabilities of the image pixels belonging to instance number $j$, while $s_{V,i}$ is a value within $[0, 1]$.

Since the Mask R-CNN often produces thousands of candidates in an image, only instances that have a high predicted $IoU$ of the bounding box above a threshold $\tau$ are considered as valid instances. After thresholding, each pixel is usually assigned to an instance number by performing the $\argmax$ per pixel to obtain the highest probability for an instance number. However, this procedure might lead to problems, as will be explained in the next paragraph.

\subsection{Post-processing}
\label{subsec:post_processing}
One of the peculiarities in the endoscopic instrument segmentation occurs in cases where instruments overlap or are only partially visible on the side of an image. In both cases the bounding box proposals may overlap, with all containing a high score which can make it difficult to proceed. These bounding box ambiguities~\cite{robu2020towards} can, when the usual pixel-wise $\argmax$ operation is performed on the results, lead to a segmentation with mixed and distributed instance numbers.

\paragraph{Remove overlapping instances}
As usual, when working with Mask R-CNN, during the first step all predicted instances that have a score greater than $\tau$ are considered as valid instances $J_V^{valid}:= \{j_{V,i} | (j_{V,i}, s_{V,i}) \in J_{V} \wedge s_{V,i} > \tau\}$. In the second step, however, all valid instances are pairwise compared with the $DSC$ metric to find proposals that have a large overlap  $J^{filtered}_V = \{ j_{V,i} \in J^{valid}_V | \nexists j_{V,\tilde{i}} \in J^{valid}_V : DSC(j_{V,i}, j_{V,\tilde{i}}) \geq \gamma \wedge |j_{V,\tilde{i}}| > |j_{V,i}|\}$ (where $|j|$ denotes the number of pixels in the instance $j$). In every case, where the instances have a high degree of overlap ($DSC\geq\gamma$), the proposal with fewer pixels is removed.

\begin{figure}[htb]
    \centering
    \begin{tabular}{p{0.4\linewidth} p{0.4\linewidth}}
    \includegraphics[trim=0 0 0 0, clip, width=1\linewidth]{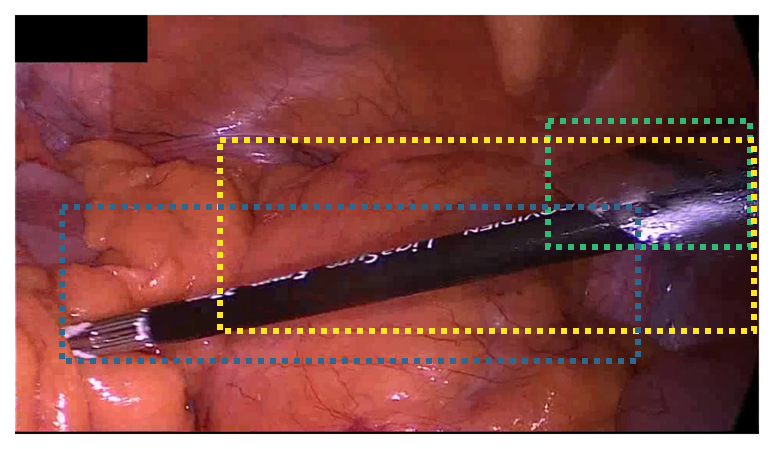}&
    \includegraphics[trim=0 0 0 0, clip, width=1\linewidth]{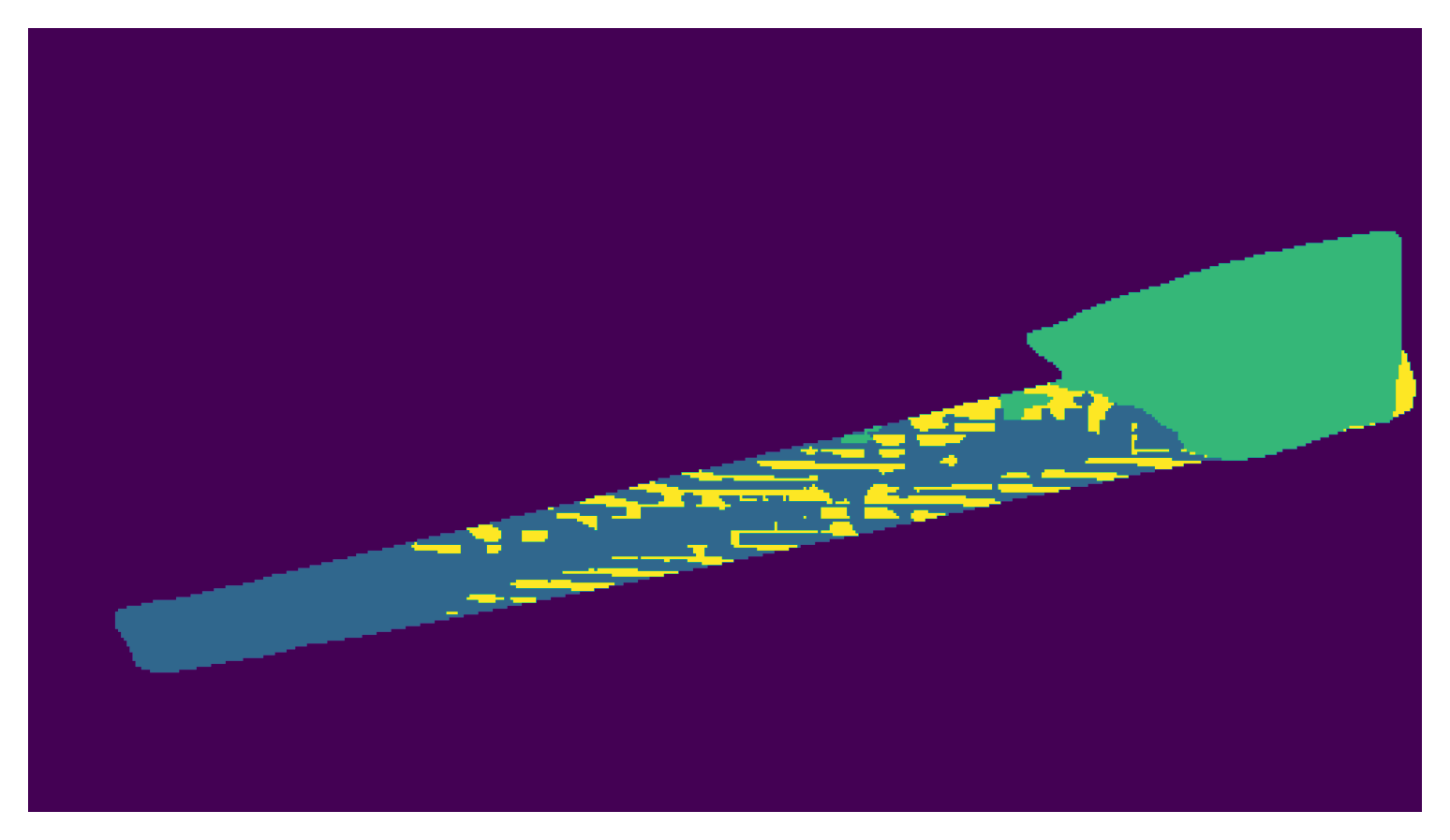}\\
    a) Image with bounding box proposals& 
    b) Segmentation before post-processing\\
    \multicolumn{2}{c}{\includegraphics[trim=0 0 0 0, clip, width=0.4\linewidth]{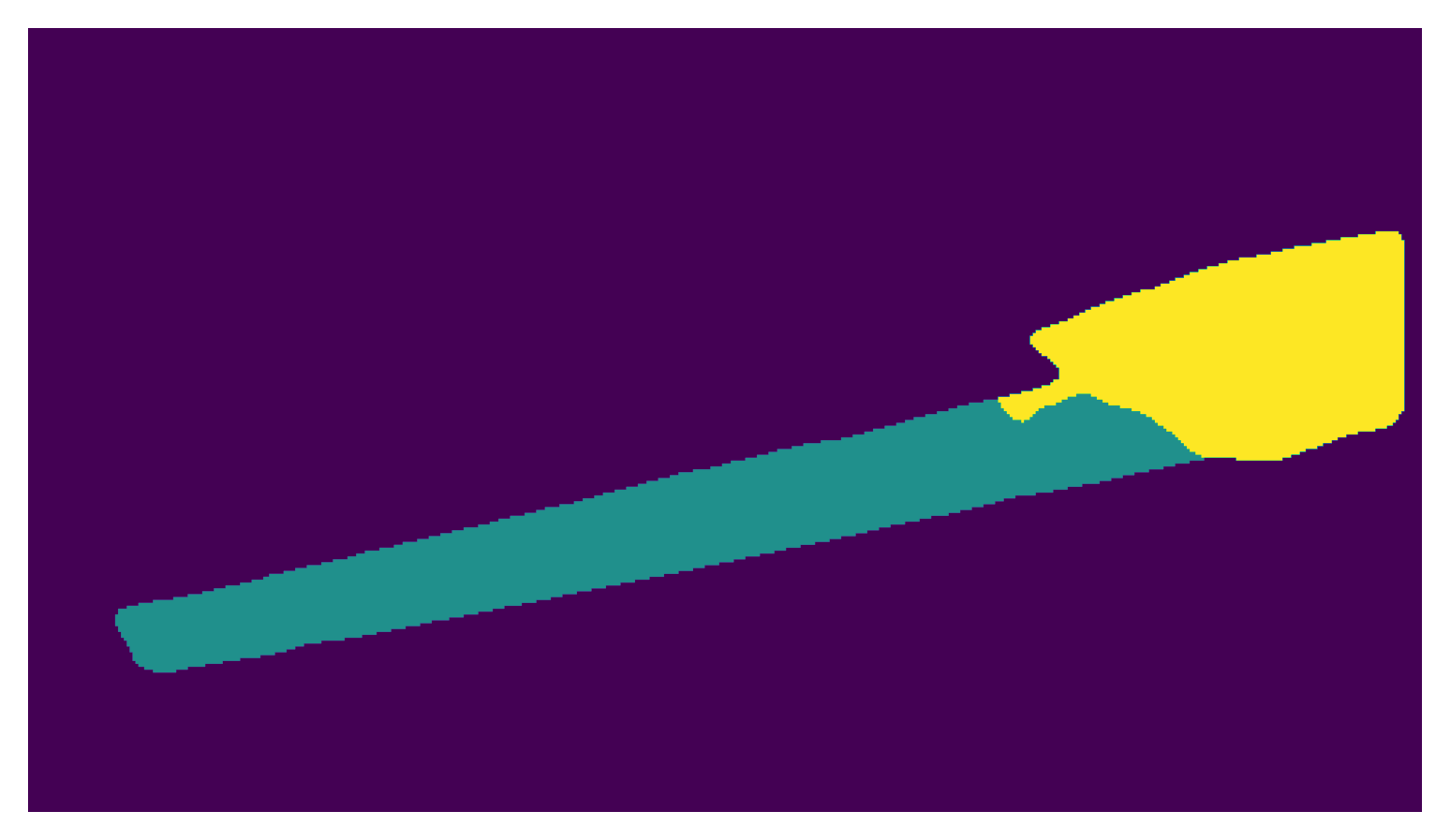}}
    \\
    \multicolumn{2}{c}{c) Segmentation after post-processing}
    \\
    \end{tabular}

    \caption[]{Effect of post-processing.  (a) Original RGB image with the detected bounding boxes and algorithm output (b) before and (c) after post-processing.}
    \label{fig:post_processing}
\end{figure}

\paragraph{Refine labels}
After multiple proposals for the same instance were removed, there remains the possibility of overlapping bounding boxes, especially in the case where instruments overlap (see Fig.~\ref{fig:post_processing}).\\

For this purpose, a Conditional Random Field~\cite{lafferty2001conditional} is used, where the unary energy is defined as the probability of each pixel belonging to an instance. After fitting the CRF, the $\argmax$ operation is applied on each pixel to assign the corresponding instrument instance number. In a final check, the result of the $\argmax$ is binarized and compared to the binary segmentation $l_{x_t}$ from the instrument likelihood prediction. If only small regions of pixels are left (number of pixels $\leq \delta$), a connected component analysis~\cite{shapiro1996connected} is performed to assign those small regions to the labels previously assigned with the CRF. If there are still large areas of instruments remaining, those are added as an additional class to the CRF and the CRF is fitted again to include the missing pixels. The results of the post-processing can be seen in Fig.~\ref{fig:post_processing}

\subsection{Implementation details}
All parameters were set according to small experiments on a held-out validation set during the implementation process.

\paragraph{Instrument likelihood prediction}
The instrument likelihood prediction is based on the semantic instrument segmentation method of Isensee et al.~\cite{isensee2020or} and thus uses an ensemble vote of eight models that were trained by omitting one surgery on an 8-fold cross-validation. Each model was trained for $2.000$ epochs using randomly selected patches at a size of $256x448$, where an epoch was defined as an iteration of over 100 batches. To prevent overfitting, the following data augmentation techniques were randomly applied on each image during training: contrast, brightness and gamma-augmentation, image rotation, scaling, mirroring, elastic deformation and additive Gaussian noise~\cite{isensee2020or}. The optimizer was the SGD~\cite{kiefer1952stochastic}, with an initial learning rate of $1$ which decayed to $0$ over the course of the training~\cite{isensee2020or}. 

The prediction model $\mathcal{B}$ was trained with the following loss $\mathcal{L}_{Seg}=\mathcal{L}_{DSC} + \mathcal{L}_{CE}$, a combination of the regular pixel-wise cross-entropy loss $\mathcal{L}_{CE}$ (Eq. \ref{eq:crossentropy}) and the soft dice loss $\mathcal{L}_{DSC}$ (Eq. \ref{eq:dsc_loss}).


%
\begin{equation}
\mathscr{L}_{CE}=-\sum y_i \log(p_i) 
\label{eq:crossentropy}
\end{equation}
\begin{equation}
\mathscr{L}_{DSC}=-\frac{2TP+ \epsilon}{2TP+FP+FN + \epsilon}
\label{eq:dsc_loss}
\end{equation}
Where $y_i$ is the reference value and $p_i$ is the estimated probability value for a given pixel $i$, $TP$ are soft true positives, $FN$ are soft false negatives and $FP$ are soft false positives).

\paragraph{Image flow prediction}
For all experiments, the optical flow was computed via a video sequence that lasted $5s$. Since using the original video frame rate of $25$ fps would result in a vanishing gradient and require too much memory, the optical flow was computed with lowered frame rate of only $3$ fps. This resulted in an optical flow with the resolution of $[15 \times h \times w \times 2]$. As the computation of the optical flow is resource-intensive, it was pre-computed and stored.\\

Summarized was the optical flow with the LSTM using $6$ hidden convolutional layers of dimension $(32, 16, 8, 4, 2, 1)$. The optical flow consists of two components (x / y component), each being summarized on its own, but all components with the same model.

\subsection{Multi-instance segmentation}
The Mask R-CNN is based on a ResNet-50 backbone, with the anchor sizes of $(8, 16, 32, 64, 128, 256, 512, 1024)$ and aspect ratios of $(0.5, 1.0, 2.0)$. The detection threshold $\tau$ was $0.2$ in order to also detect small instrument boundaries. For training the model, the SGD optimizer was used, with a momentum of $0.9$, a learning rate of $0.01$ and a batch consisting of $2$ samples. To prevent the training from crashing, all gradients were clipped at $1.0$. Since the meaning of the optical flow is not invariant to augmentation techniques that change pixel locations, no data augmentation techniques were used. Only the image dimension was halved, with respect to the optical flow vector.

\subsection{Post-processing}
For the post-processing, $\gamma$ (overlap between proposals) was set to $0.5$ and $\delta$ (very small pixel regions) was set to $100$. The parameters were estimated on a small validation set that was randomly taken from the training data set.

\end{document}